\pdfoutput=1

\documentclass[11pt]{article}

\usepackage[preprint]{acl}

\usepackage{times}
\usepackage{latexsym}
\usepackage{booktabs}
\usepackage{tipa}
\usepackage{multirow}

\usepackage[T1]{fontenc}

\usepackage[utf8]{inputenc}

\usepackage{microtype}

\usepackage{inconsolata}

\usepackage{graphicx}

\title{Annotating and Inferring Compositional Structures in Numeral Systems Across Languages}

\author{
  \textbf{Arne Rubehn\footnotemark[1]\textsuperscript{1}},
  \textbf{Christoph Rzymski\thanks{Equal contribution}\textsuperscript{2}},
  \textbf{Luca Ciucci\textsuperscript{1}},
  \textbf{Kellen Parker van Dam\textsuperscript{1}}, \\
  \textbf{Al{\v{z}}b{\v{e}}ta Ku{\v{c}}erová\textsuperscript{1}},
  \textbf{Katja Bocklage},
  \textbf{David Snee\textsuperscript{1}},
  \textbf{Abishek Stephen\textsuperscript{3}},
  \textbf{Johann-Mattis List\textsuperscript{1}}
\\
\\ 
  \textsuperscript{1}Chair for Multilingual Computational Linguistics, University of Passau, Passau, Germany \\
    \textsuperscript{2}Department of Linguistic and Cultural Evolution, Max Planck Institute for \\ Evolutionary Anthropology, Leipzig, Germany \\
  \textsuperscript{3}Institute of Formal and Applied Linguistics, Charles University, Prague, Czech Republic
}

\begin{document}
\maketitle
\begin{abstract}
Numeral systems across the world's languages vary in fascinating ways, both regarding their synchronic structure and the diachronic processes that determined how they evolved in their current shape. For a proper comparison of numeral systems across different languages, however, it is important to code them in a standardized form that allows for the comparison of basic properties. Here, we present a simple but effective coding scheme for numeral annotation, along with a workflow that helps to code numeral systems in a computer-assisted manner, providing sample data for numerals from 1 to 40 in 25 typologically diverse languages. We perform a thorough analysis of the sample, focusing on the systematic comparison between the underlying and the surface morphological structure. We further experiment with automated models for morpheme segmentation, where we find allomorphy as the major reason for segmentation errors. Finally, we show that subword tokenization algorithms are not viable for discovering morphemes in low-resource scenarios.

\end{abstract}

\section{Introduction}

Numeral systems represented by the words for cardinal numbers used in counting are an interesting kind of linguistic data: they code a part of the lexicon of human languages that is potentially large and often exhibits a regularity that increases with higher numbers. Regularity is reflected in the \emph{recycling} of linguistic material used to create higher numbers, where morphemes for smaller number words are often reused to motivate the formation of larger numerals. In addition, numeral systems are also maximally \emph{distinctive}. Being used to distinguish ordinal numbers, we rarely find cases in which two distinct numbers are expressed by the same word form, even if numeral words themselves can have multiple meanings outside of the number domain (as can be seen in the \emph{Database of Cross-Linguistic Colexifications}, \citealp{Rzymski2020}).

Another important aspect of numeral systems is that they are not created in an ad-hoc fashion but have instead often evolved over hundreds of years. The evolution can leave traces in numeral systems that counter-act former regularity, leading to allophonic variation in the morphemes that compose numeral words. Language contact can also feature as an important aspect of evolution, resulting in extreme cases where languages use two or more numeral systems in combination, reflecting different stages of their history.

The fact that most numeral systems are \emph{compositional}, while at the same time being distinctive and discrete in their denotation, makes them an interesting test object for linguistic analyses that deal with lexical compositionality in the context of language change. While one would otherwise have to cope with problems resulting from various kinds of morphological and semantic variation, numeral systems can be seen as an ideal test ground for the annotation and inference of compositional structures in the lexicon of human languages. In the following, we will try to illustrate this point in more detail. After a short overview on numeral systems in the context of descriptive and computational linguistics (§~\ref{sec:2}), we present a small collection of numeral systems along with methods that can be used to annotate numeral systems manually or to segment numeral words automatically into morphemes (§~\ref{sec:3}). After testing these methods and reporting the results on our small cross-linguistic sample of numeral systems (§~\ref{sec:4}) we discuss our findings and point to ideas for future work (§~\ref{sec:5}).


\section{Background}\label{sec:2}

The cross-linguistic diversity of numeral systems has attracted the interest of scholars since \citeauthor{HervasyPanduro1786}'s comparative work (\citeyear{HervasyPanduro1786}), which presented data from missionaries on many then little-known languages. Today, the most comprehensive database on numerals is \citet{Chan2024}, who collected data on more than 5,000 lects, often provided by linguists with first-hand experience of the respective languages. The constant increase in data has allowed for the study of numeral systems from a formal \citep[see e.g.][]{Corstius1968,Hurford1975} and a typological perspective. The latter approach reached a turning point with \citeauthor{Greenberg1978}'s (\citeyear{Greenberg1978}) 54 generalizations, most of which stood the test of time \citep{Comrie2020}.


Even though their synchronic structure may be opaque, numeral systems are diachronically motivated and are built through a limited number of cross-linguistic strategies \citep[18-34]{Heine1997}. They typically combine a small set of morphemes (mainly numbers, but also linking elements) according to three parameters, including (1)~the choice of the base(s), (2)~the operations applied to the base(s), 
and (3)~the order of the morphemes \citep[459-461]{Greenberg1978,Moravcsik2017}. Despite the presumed regularity and compositionality of numeral systems, they may occasionally display gaps and ambiguities \citep[79-80]{Comrie1997,Comrie2005}.

The most common bases are `five', `ten', and `twenty', whose conceptual sources are, respectively, the fingers of the hand, of both hands, and of all hands and toes (\citealp[19-24]{Heine1997}; on finger counting and its cultural variability, see \citealp{Bender2012}). 
Decimal systems are the most frequent worldwide, followed by vigesimal and quinary systems (\citealp{Skirgard2023}). Languages can employ more than one base, resulting in hybrid
numeral systems.

While languages with no numerals or only the number `one' are rare \citep{Hammarstrom2010}, the numeral systems of many languages, particularly in South America, New Guinea and Australia, are restricted to a few numerals \citep[459]{Moravcsik2017}. According to \citet[71-72]{Dixon2012}, this indicates that the speakers did not count and enumeration was not the primary use of these number words. 
\citet{hammarstrom2008complexity} observed that pidgins and creoles tend to have more complex numeral systems than the global average. Their frequent origin as trade languages may be a contributing factor. Numeral systems often developed out of contact, which usually comes with societal change, and borrowing may also involve the lowest numbers \citep[75-77]{Dixon2012}. 

While numeral systems all over the world have been quite intensively investigated in the past, formal computational studies that account for the degree of compositionality and the individual motivation patterns underlying individual number words have not been carried out so far. Recent advances in the annotation of lexical motivation patterns \citep{Hill2017a} and the automated segmentation of words into morphemes \citep{goldsmith2017computational} open new possibilities for a computational investigation of numeral systems that we will discuss in more detail in the following.

\section{Materials and Methods}\label{sec:3}

\subsection{Sample of Numeral Systems}

\begin{figure}[tb]
    \centering
    \includegraphics[width=\linewidth]{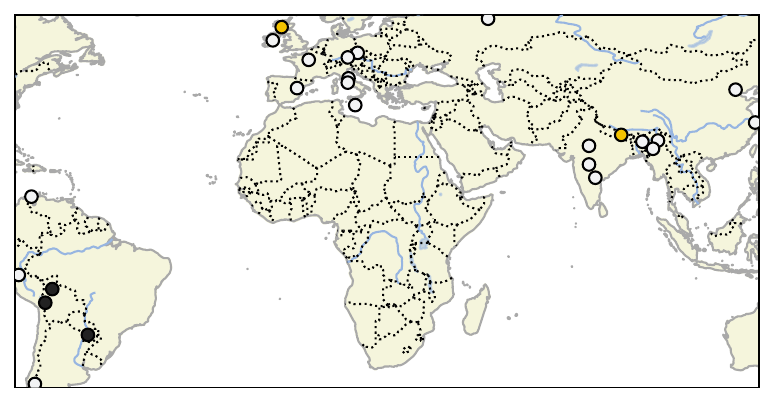}
    \caption{Geographical distribution of the languages in our sample, indicating the numeral bases they employ (white: 10, black: 5 and 10, orange: 10 and 20).}
    \label{fig:map}
\end{figure}

We collected the cardinal numbers from 1 to 40 in 25 typologically diverse languages from Eurasia and Southern America, spanning ten different language families. Most language families, with the exception of Indo-European (12 languages) and Sino-Tibetan (5 languages), are represented by a single language. Table \ref{tab:overview} provides a comprehensive overview of the languages covered in the sample, accompanied by a geographical visualization in Figure \ref{fig:map}.

\begin{table}[tb]
\centering
\resizebox{\linewidth}{!}{%
\begin{tabular}{|l|l|l|l|}
\hline

\textbf{Family}        & \textbf{Branch}         & \textbf{Language}  & \textbf{Base} \\ \hline
Afro-Asiatic  & Semitic        & \href{https://glottolog.org/resource/languoid/id/malt1254}{Maltese}   & 10   \\\hline
Araucanian    & ---            & \href{https://glottolog.org/resource/languoid/id/mapu1245}{Mapudungun}   & 10   \\\hline
Arawak        & Ta-Arawak      & \href{https://glottolog.org/resource/languoid/id/wayu1243}{Wayuu}   & 10   \\\hline
Aymaran       & ---            & \href{https://glottolog.org/resource/languoid/id/ayma1253}{Aymara}   & 5 / 10   \\\hline
Dravidian     & Southern       & \href{https://glottolog.org/resource/languoid/id/telu1262}{Telugu}   & 10   \\ \hline
\multirow{11}{*}{Indo-European} & \multirow{2}{*}{Balto-Slavic}   & \href{https://glottolog.org/resource/languoid/id/czec1258}{Czech}   & 10     \\ \cline{3-4}
              &                & \href{https://glottolog.org/resource/languoid/id/russ1263}{Russian}   & 10   \\\cline{2-4}
              & \multirow{2}{*}{Celtic}         & \href{https://glottolog.org/resource/languoid/id/iris1253}{Irish}   & 10   \\ \cline{3-4}
              &                & \href{https://glottolog.org/resource/languoid/id/scot1245}{Scottish Gaelic}   & 10 / 20   \\\cline{2-4}
              & Germanic       & \href{https://glottolog.org/resource/languoid/id/stan1295}{German}   & 10   \\ \cline{2-4}
              & \multirow{3}{*}{Indo-Iranian}   & \href{https://glottolog.org/resource/languoid/id/assa1263}{Assamese}   & 10   \\ \cline{3-4}
              &                & \href{https://glottolog.org/resource/languoid/id/hind1269}{Hindi}   & 10   \\ \cline{3-4}
              &                & \href{https://glottolog.org/resource/languoid/id/sans1269}{Sanskrit}   & 10   \\ \cline{2-4}
              & \multirow{4}{*}{Romance}         & \href{https://glottolog.org/resource/languoid/id/stan1290}{French}   & 10   \\ \cline{3-4}
              &                & \href{https://glottolog.org/resource/languoid/id/ital1282}{Italian}   & 10   \\ \cline{3-4}
              &                & \href{https://glottolog.org/resource/languoid/id/lati1261}{Latin}   & 10   \\ \cline{3-4}
              &                & \href{https://glottolog.org/resource/languoid/id/stan1288}{Spanish}   & 10   \\\hline
Pano-Takanan  & Takanan        & \href{https://glottolog.org/resource/languoid/id/cavi1250}{Cavineña}   & 5 / 10   \\\hline
Quechuan      & Quechua I      & \href{https://glottolog.org/resource/languoid/id/hual1241}{Huallaga Quechua}   & 10   \\\hline
\multirow{4}{*}{Sino-Tibetan}  & Bodic          & \href{https://glottolog.org/resource/languoid/id/lamj1247}{Lamjung Yolmo}   & 10 / 20   \\ \cline{2-4}
              & Brahmaputran   & \href{https://glottolog.org/resource/languoid/id/khoi1251}{Uipo (Maringic)}   & 10   \\ \cline{2-4}
              & Patkaian & \href{https://glottolog.org/resource/languoid/id/maky1236}{Makyam}   & 10   \\\cline{2-4}
              & \multirow{2}{*}{Sinitic}        & \href{https://glottolog.org/resource/languoid/id/mand1415}{Mandarin Chinese}   & 10   \\\cline{3-4}
              &                & \href{https://glottolog.org/resource/languoid/id/shan1293}{Shanghainese}   & 10   \\\hline
Tupian        & Tupí-Guaraní   & \href{https://glottolog.org/resource/languoid/id/para1311}{Paraguayan Guaraní}   & 5 / 10   \\ \hline
\end{tabular}}
\caption{Overview of languages covered in the sample, with their genetic classification and primary bases for counting.}
\label{tab:overview}
\end{table}

Most languages employ a decimal system, reflecting that the number 10 is by far the most common base. Three languages in our sample -- Aymara, Cavine\~{n}a, and Paraguayan Guaraní -- make use of the number 5 as a base. They represent a hybrid between quinal and decimal systems, since the word for 10 is monomorphemic and used to express multiples of 10. Furthermore, two languages of our sample (Lamjung Yolmo and Scottish Gaelic) have retained a vigesimal system used in parallel to a decimal system, which results in alternating forms for numbers higher than 20.

All data were collected, annotated, and curated in a collaborative manner, such that the data for each language were thoroughly reviewed by at least two scholars: the responsible annotator for the given language, and at least one reviewer. The data were then aggregated and deployed as a unified dataset conforming to the Cross-Linguistic Data Formats \citep[CLDF,][]{forkel2018cross,Forkel2020}. Automated tests accounted for the structural integrity of the data (e.g. ensuring that one cognate ID does not map to more than one underlying form; the annotation format is described in detail in § \ref{sec:annot}).


\subsection{Representing Numeral Systems in Tables}
The CLDF specification builds on CSVW, a standard for tabular data on the web \citep[\href{https://csvw.org}{https://csvw.org};][]{CSVW} that extends simple tabular data, typically represented in the form of CSV files, by metadata that can be used to specify the content of tabular data in various ways, including the combination of multiple tables in a relational database. Given that numeral data can be easily treated as \emph{lexical data}, typically provided in the form of wordlists, we represent number systems as extended CLDF wordlists that build on the extended wordlist formats introduced by the Lexibank repository \citep{List2022e}. Lexibank wordlists represent individual word forms as triples consisting of a \emph{language}, a \emph{concept}, and a \emph{form}. In order to compare data from different sources, Lexibank makes use of reference catalogs that link language varieties to Glottolog \citep[\href{https://glottolog.org}{https://glottolog.org;}][]{Glottolog}, map concepts to Concepticon \citep[\href{https://concepticon.clld.org}{https://concepticon.clld.org;}][]{Concepticon}, and represent phonetic transcriptions compatible with the subset of the IPA proposed by the Cross-Linguistic Transcription Systems (CLTS) reference catalog \citep[\href{https://clts.clld.org}{https://clts.clld.org;}][]{CLTS}. 

While following Lexibank in assembling our exploratory database of numeral systems, we extend the format by adding new layers of annotation that help us to make individual analyses of the numeral systems explicit through annotation. As a first step, we rigorously split words into morphemes by adding morpheme boundary markers to all multi-morphemic words (using the plus symbol -- \texttt{+} -- as a boundary marker). As a second step, we identify language-internal partial cognates in all numeral systems in order to mark the degree by which morphemes are reused to build new numeral expressions (see \citealp{List2016g} on partial cognates). As a third step of analysis, we add \emph{morpheme glosses} to the data to add human-readable semantic hints to all morphemes \citep{Hill2017a,Schweikhard2020}. As a fourth step, we make use of \emph{inline-alignments} in order to handle allomorphs by distinguishing underlying from surface forms \citep{Pulini2024,List2024fc}. As a fifth step, we conduct \emph{phonetic alignment analyses} \citep{List2014d} of all language-internal cognate morphemes, in order to facilitate the comparison of allomorphic variants that differ in length.

\begin{table*}[tb]
\centering
\resizebox{0.9\textwidth}{!}{%
\tabular{|l|l|l|l|l|}
\hline
\bfseries Language &
\bfseries Concept & 
\bfseries Segments &
\bfseries Cognates &
\bfseries Morphemes \\\hline\hline

German & one        & \textipa{aI n s/-}                                 & 1 & \ttfamily ONE  \\\hline
German & two        & \textipa{ts v aI}                                & 2 & \ttfamily TWO  \\\hline
German & three      & \textipa{d r aI}                                 & 3 & \ttfamily THREE \\\hline
German & twenty one & \textipa{ai n s/-} + \textipa{U n -/d} + \textipa{ts v a n} + \textipa{ts I \c{c}} & 1 4 5 6  & \ttfamily ONE and TWEN TY \\\hline 
German & thirty two & \textipa{ts v aI} + \textipa{U n -/d} + \textipa{d r aI} + \textipa{s/ts I \c{c}}  & 2 4 3 6  & \ttfamily THREE and THREE TY \\\hline \hline
French & one & \textipa{\~\oe } & 1& \ttfamily ONE \\\hline
French & two & \textipa{d \o} & 2& \ttfamily TWO \\\hline
French & three & \textipa{t K w a} & 3& \ttfamily THREE \\\hline
French & twenty one & \textipa{v \~E t/-} + \textipa{e} + \textipa{\~\oe} & 4 5 1& \ttfamily TWENTY and ONE\\\hline
French & thirty two & \textipa{t K -/w -/A} + \textipa{\~A t} + \textipa{d \o}& 3 6 2& \ttfamily THREE TY TWO \\\hline
\endtabular
}
\caption{Illustration of the format used to annotate morpheme boundaries along with allomorphic variation, language internal cognates, and morpheme glosses.}
\label{tab:ann}

\end{table*}

Table \ref{tab:ann} shows how our annotations are rendered in tabular form, with examples for annotated numerals from German and French. The column \emph{Segments} provides phonetic transcriptions, segmented into sounds, using a space as boundary marker, and secondarily segmented into morphemes, using the plus symbol as a boundary marker. The transcriptions use \emph{inline alignments} \citep{List2024fc} to align the surface forms with their underlying forms. Inline alignments use the slash symbol (\texttt{/}) in order to distinguish a surface sound (shown to the left of the slash) from its corresponding underlying sound. As an example, consider the transcription of German \textipa{[aI n s/-]} `one' in the table, where the sound \textipa{[s]} is treated as a surface form, while the underlying form does not show this sound (which is marked by using the gap-symbol \texttt{-} after the slash). The notion of surface form and underlying form is strictly technical. We assume that one morpheme with multiple allomorphs has only one underlying form, which must consistently be aligned with all surface forms. We do not claim that this handling shows any cognitive truth, but we aim for an annotation that would ideally be meaningful from a diachronic perspective.

The columns \emph{Cognates} and \emph{Morphemes} provide information on language-internal cognates in the form of morphemes that are reused. Here, the \emph{Cognates} column employs numerical identifiers, following the format proposed by \citet{List2016g}, while the functionally identical \emph{Morphemes} column provides semantic glosses that help in making the lexical motivation underlying the formation of numerals transparent. This annotation, which provides explicit glosses for all morphemes constituting a word, was originally developed to make language-internal cognate relations more explicit \citep{Hill2017a}. By now, however, it has been shown to be also very useful to provide rudimentary annotations of lexical motivation patterns \citep{Brid2022b}.

\subsection{Computer-Assisted Annotation}\label{sec:annot}

While the annotations shown in Table \ref{tab:ann} can be easily carried out with the help of a spreadsheet editor or directly in text files, we use the web-based EDICTOR tool for the annotation of numeral data \citep{EDICTOR}. Originally, EDICTOR was designed to facilitate the process of creating multilingual comparative wordlists \citep{List2017d}. Since Version 3.0 \citep{List2024b}, however, EDICTOR has been substantially extended to help with the annotation of lexical motivation patterns. Improvements include -- among others -- a visual rendering of inline alignments, sound sequences, cognate sets, and morpheme glosses, combined with annotation helpers for manual morpheme segmentation, as well as several sanity checks that increase the consistency of human annotation.

\subsection{Automated Morpheme Segmentation}\label{sec:morseg-models}

The task of unsupervised morpheme segmentation -- automatically inferring a language's morphological structure from unannotated corpus data -- has received notable attention in the field of Natural Language Processing, especially in the late 1990's and early 2000's \citep{hammarstrom2011ulm}. While those models were developed with a different background in mind, assuming the presence of relatively large training corpora, numeral systems naturally lend themselves as an interesting use case for morpheme segmentation models due to their high degree of compositionality. Therefore, we experiment with simple morpheme segmentation techniques to observe their performance in a transfer setting with much less data, but an extraordinarily strong morphological signal.

The first formalization of an algorithm for morpheme segmentation reaches back to \citet{harris1955from} who proposed the so-called  \emph{Letter Successor Variety} (LSV) as a predictability measure at each position within a word. The underlying assumption is that the continuation of a word should be fairly predictable within a morpheme, but much harder to predict at a morpheme boundary. 
Several proposals have been made to improve upon LSV. \citet{hafer1974word} suggest measuring predictability in terms of entropy rather than type variety. They also propose Letter Predecessor Variety (or Entropy) as a logical inversion of LSV, processing each word backwards. \citet{hammarstrom2009unsupervised} proposes \emph{Letter Successor Max-Drop}, measuring how likely the most frequent continuation of a word is in comparison to all other potential continuations. 
 We experiment with all these different flavors of LSV, but report only LSE, since it performs best on average and all LSV variations show similar patterns in general. Following \citet{hafer1974word}, we also experiment with a simple model that considers every possible prefix and suffix (in a computational sense) of a word form as a morpheme if and only if it appears as a complete word in the data. Using this simple measure, \citet{List2023a} reports promising results in inferring partial colexifications from multilingual wordlists which seem to advance concept embeddings substantially \citep{Rubehn2025PREPRINT}.

A line of research that can be seen as complementary to LSV-based approaches formalizes the task of morpheme segmentation as a \textit{minimum description length} (MDL) problem \citep{goldsmith2001unsupervised}. The basic idea behind MDL is to define a description length as a combination of basic tokens and rules to derive complex forms from the basic vocabulary. This notion is especially interesting on theoretical grounds, since the complexity of numeral systems can also be measured in terms of MDL \citep{hammarstrom2008complexity}. In an ideal setting, an MDL-based segmentation model is therefore expected to accurately infer and model the compositional structure of numeral systems. Representing this family of morpheme segmentation algorithms, we run our experiments with the Morfessor Baseline model \citep{creutz2002unsupervised,creutz2005unsupervised,virpioja2013morfessor}.

\subsection{Subword Tokenization}\label{sec:subword-models}

Algorithms for \textit{subword tokenization} form an integral preprocessing step of state-of-the-art language models, since they effectively reduce the vocabulary size and avoid the occurrence of out-of-vocabulary items. While these tokenization methods in principle make downstream applications more flexible, it can at least be doubted whether the inferred subwords concord with the language's morphological structure \citep{batsuren2024evaluating}. We apply three popular algorithms for subword tokenization on our multilingual numeral data: Byte-Pair-Encoding \citep[BPE; ][]{gage1994new,sennrich2016neural}, WordPiece \citep{schuster2012japanese}, and Unigram tokenization \citep{kudo2018subword}.

\subsection{Evaluation}

All models described in §~\ref{sec:morseg-models} and §~\ref{sec:subword-models} are trained on unannotated and unsegmented representations of the numeral lists. The predicted segmentations are then evaluated against our manual annotations which serve as a gold standard. Since all models are inherently monolingual, each language is processed and evaluated independently.

Predicted segmentations can directly be evaluated against the gold standard using \textit{precision} and \textit{recall} \citep{virpioja2011empirical}. While we are aware of more sophisticated evaluation metrics for morphological analyses \citep{spiegler2010emma}, we argue that simply calculating 
boundary precision and recall (BPR) is sufficient in our use case, since we investigate small corpora with hardly ambiguous morphological patterns. Due to its simplicity, BPR is readily interpretable, rendering it the ideal evaluation metric for our use case.

We run all experiments on two different representations of the numeral lists, relying on the \textit{surface} and \textit{underlying} forms respectively (see §~\ref{sec:annot} for details on the two representations). 
Comparing these two settings allows for a fine-grained evaluation of morpheme segmentation models, enabling us to assess the share of segmentation errors caused by allomorphy.

\subsection{Implementation}

The data were annotated using EDICTOR 3.1 \citep{EDICTOR}, and validated and compiled using CLDFBench \citep{Forkel2020}. The visualization in Figure \ref{fig:map} was created using CLDFViz \citep{CLDFViz}. All experiments regarding automated morpheme segmentation were run in Python, using LinSe \citep{forkel2024linse} to conveniently represent the internal structure of word forms in different granularities. Morfessor was run from its Python package \citep{virpioja2013morfessor}, all other models were implemented from scratch. All data and code accompanying this study are made available in the supplementary material.

\section{Analysis and Results}\label{sec:4}

\subsection{Sample Data of Coded Numeral Systems}

\begin{table}[t]
\resizebox{\linewidth}{!}{%
\begin{tabular}{@{}|l|cc|cc|cc|@{}}
\hline
& \multicolumn{2}{c|}{\textbf{Average}}        & 
\multicolumn{2}{c|}{\textbf{Highest}}   & 
\multicolumn{2}{c|}{\textbf{Lowest}} \\
& \textbf{S} & \textbf{U} & \textbf{S} & \textbf{U} & \textbf{S} & \textbf{U} \\\hline
\textbf{Morphemes}    & 21.8 & 13.5 & 48 & 20   & 10 & 7    \\
\textbf{Expressivity} & 5.6 & 7.9   & 10.6 & 15 & 1.4 & 3.4 \\
\textbf{Opacity}      & \multicolumn{2}{c|}{1.60}        & \multicolumn{2}{c|}{3.18}      & \multicolumn{2}{c|}{1}         \\
\textbf{Code Length}  & \multicolumn{2}{c|}{2.53}        & \multicolumn{2}{c|}{3.83}      & \multicolumn{2}{c|}{11.68}      \\ \hline
\end{tabular}}
\caption{Overview of statistics about the different numeral systems. \textbf{S} and \textbf{U} refer -- where applicable -- to surface vs. underlying forms.}
\label{tab:stats}
\end{table}

Table \ref{tab:stats} summarizes the results of computing different types of metrics based on surface and underlying forms across all languages in our sample (Table \ref{tab:stats-all-langs} in the appendix provides metrics for individual languages).
In the table, we introduce three simple metrics -- \emph{expressivity}, \emph{opacity}, and \emph{length} -- to get a better understanding of the data and the strategies to form higher numbers from basic morphemes. First, we measure the average \textit{morpheme expressivity} of a language by counting how many different numbers are formed using this morpheme. For the rare cases where a language has multiple forms for the same number, expressivity is weighted accordingly.
\textit{Opacity} describes the ratio between allomorphic variants and morphemes, measuring the degree of allomorphy in a system. The lowest score is 1, with each morpheme in a language surfacing with the same form. Finally, the \textit{average coding length} measures how many morphemes are used to form a word.

On theoretical grounds, the minimum amount of morphemes required in a numeral system is the base of that system. That means, a decimal system needs at least 10 different morphemes to be fully expressive. Indeed, our sample covers three languages -- Mandarin, Mapudungun, and Huallaga Quechua -- that use such a minimal decimal system to express the numbers up to 40. This observation holds true on both the surface and the underlying level, indicating that exactly these languages lack any kind of allomorphy. On the other side of the spectrum, we find Assamese with 20 different morphemes and Hindi with 48 distinct morphs, the highest value for the respective category. This aligns with the general impression that Indo-Aryan languages feature some of the most complex and opaque numeral systems of the world \citep{hammarstrom2008complexity}.

We observe a wide range of morpheme opacity. With Uipo, Huallaga Quechua, Mandarin, and Mapudungun, four languages in our sample have the lowest possible opacity of 1.0, thus not featuring any allomorphy in their numeral systems. On the other hand, we find Lamjung Yolmo with an opacity of 3.18, indicating that each morpheme on average is represented by a bit more than three different surface forms. Lamjung Yolmo is followed by Hindi, Telugu, and Sanskrit, three languages that are (or were) spoken in India. From these extreme cases, the impression might arise that the opacity correlates with the size of the underlying morpheme inventory. However, across the entire dataset, no significant correlation between these two metrics could be found.

The expressivity of morphemes and their allomorphic variants, on the other hand, shows a significant negative correlation with the number of morphemes. The interpretation is straightforward: The fewer morphemes are available in a system, the more expressive they need to be, and the more they will be used. It is therefore not surprising that exactly those three languages that employ a base of 5 (Aymara, Cavineña, and Paraguayan Guaraní) rank the highest in terms of expressivity on the surface and the underlying level. On the low end of expressivity, we again find Hindi and Assamese, as well as the modern Romance languages French, Italian, and Spanish.

Based on these correlations, one might expect that the average coding length is also directly dependent on the size of the morpheme inventory, since less available morphemes should -- in theory -- require longer word forms. However, no significant correlation between these two metrics could be found. There is only a significant correlation between the coding length and the morpheme expressivity. Considering that our sample is heavily biased towards decimal systems, and that even the systems that employ other bases show traces of decimal coding, we cannot interpret these effects as a result of different numeral bases. Instead, this seems to result from oblique marking (connecting morphemes with particles like `and' or `with') which can happen independently of the numeral base.

Finally, we experiment with \textit{type-token ratio} (TTE) and \textit{entropy}, which have been proposed as measures of morphological complexity in the past \citep{bentz2017entropy,coltekin2022complexity}. These metrics are not able to capture any aspect of complexity in our sample, since they correlate almost perfectly with the number of morphemes. We therefore conclude that in this special setting, TTE and entropy are dependent on the vocabulary size alone, which is probably due to the fact that morphemes in numeral systems by and large do not follow a Zipfian distribution, as is the case for words in natural language corpora.

\begin{table}[b]
\centering
\resizebox{\linewidth}{!}{%
\begin{tabular}{@{}|l|r|r|@{}}
\hline
\textbf{Model} & \textbf{Surface Forms} & \textbf{Underlying Forms}  \\\hline
Morfessor & 0.74&0.87    \\
LSPE & 0.72& 0.83   \\
Affix & 0.72&0.88   \\
\hline
\end{tabular}}
\caption{Average F$_1$ scores of morpheme segmentation algorithms.}
\label{tab:morseg-stats}
\end{table}

\subsection{Automated Morpheme Segmentation}\label{sec:morseg-discussion}

Table \ref{tab:morseg-stats} reports the overall performance of three models for automated morpheme segmentation on the individual languages, both for those cases where surface forms were passed to the algorithms, and where underlying forms were taken as the basis of analysis.
From the model family based on Letter Successor Variety, we only report Letter Successor/Predecessor Entropy, which generally performed best.

The most obvious (and unsurprising) observation is that all models perform better on the underlying form than on the surface forms. Since it is a well-known issue in the literature that automated methods are challenged by allomorphy \citep{hammarstrom2011ulm,virpioja2011empirical}, this does not seem too surprising to us. Comparing the average scores of the models, however, shows that allomorphy is the biggest source of error for the analysis on surface forms, which naturally is the common use case for those models. By extension, it does not come as a surprise that opacity significantly correlated with how well the models perform on the surface forms.

But even on the underlying forms -- an ``ideal'' scenario in which allomorphy does not exist -- there are notable differences in how well the morphological structure is detected by the models. Particularly interesting is the case of Uipo. This numeral system poses a big challenge for Morfessor and the Affix model, which both only achieve an F$_1$-score of 0.4. A closer look at the language data reveals that Uipo has a complex numeral system, in which even the numbers between 2 and 9 consist of two morphemes, a prefix and a stem. The number 6 for example is \textipa{[t\super{h} @ + r u k]}, but both morphemes are only used to form the number six (and by extension, numbers that are formed using `six'). Without any further knowledge of the language, it is very hard if not impossible to recognize the underlying compositionality. 
On the other hand, the high score of LSPE on Uipo -- which may come as a surprise -- can be described as a coincidental byproduct of the present morphophonology. As generally typical for South-East Asian languages, Uipo only allows the simple syllable structure CV(C), and each syllable in Uipo is a morpheme at the same time. Since there are more consonants than vowels, the continuation of a word is much less predictable at the start of a new syllable. LSPE can therefore accurately predict \textit{syllable} boundaries, which happen to be morpheme boundaries as well.

On the other side, Morfessor is able to perfectly predict all morpheme boundaries in four languages at the surface level (Shanghainese, Mandarin, Huallaga Quechua, Mapudungun), and in seven more languages at the underlying level. Mapudungun seems to have a particularly transparent structure, since it is the only language that all three models segment perfectly at both representation levels.
This makes Morfessor the model with the highest number of completely correct segmentations at the language level, showing that it clearly has the edge over the other two approaches tested, which is also indicated by the average performance. But even in this ``ideal'' scenario -- no allomorphy and a system that shows clear compositional structures -- Morfessor cannot accurately predict all morpheme boundaries for 14 out of 25 languages. For example, in the German words \textit{zwan-zig} `twen-ty' and \textit{drei-ßig} `thir-ty', the model fails to detect the morpheme boundaries, even in the underlying form where \textit{-zig} \textipa{[ts I \c{c}]} and \textit{-ßig} \textipa{[s I \c{c}]} are represented in the same way (\{\textipa{ts I \c{c}}\}). Generally, the model is much more prone to undersplitting than to oversplitting: On the underlying representation, it achieves a nearly perfect precision of 0.998, but a recall of only 0.80.


\subsection{Subword Tokenization}

\begin{table}[b]
\centering
\resizebox{\linewidth}{!}{%
\begin{tabular}{@{}|l|r|r|@{}}
\hline
\textbf{Model} & \textbf{Surface Forms} & \textbf{Underlying Forms} \\\hline
BPE       & 0.51    & 0.61      \\
WordPiece & 0.36    & 0.35       \\
Unigram   & 0.32    & 0.32       \\
\hline
\end{tabular}}
\caption{Average F$_1$ scores of subword tokenization algorithms for morphological segmentation.}
\label{tab:subword-stats}
\end{table}

Table \ref{tab:subword-stats} provides an overview of how accurately algorithms for subword tokenization can capture the morphological structure of the numeral systems at hand. It is evident that these models are in no way competitive with algorithms designed for the task of morphological segmentation -- even the simplest segmentation algorithms outperform the subword algorithms largely. Among the subword tokenization algorithms, BPE performed the best on both levels, and the Unigram model performed worst across the board.

There are two major conceptual issues that inhibit a successful transfer of these algorithms to morpheme segmentation. First, these models only operate extremely locally -- BPE and WordPiece merge bigrams based on a simple co-occurence metric, and Unigram removes unlikely \emph{n}-grams under the assumption that the distribution of all tokens in the vocabulary is statistically independent. This prevents the models from learning relevant information about longer shared substrings, which is the foundation for all successful morpheme segmentation models.
The second, and arguably strongest limiting factor is that it is unclear how to determine when a model should stop. In their intended setting, subword tokenization algorithms are designed to define an expressive vocabulary of a tractable size for downstream NLP applications. Hence, a desired vocabulary size is defined a priori, and the subword vocabulary is continually modified until the predefined size is reached. For BPE and WordPiece, the vocabulary size increases monotonically during that process, while it decreases for Unigram.

This training set-up leads to two problems. The first is that the desired vocabulary size must be defined before running the model. For morphological segmentation, the ideal vocabulary size naturally will be the size of the morpheme inventory -- but if that is already known, then no automated morphological analysis is required anymore. 
For the sake of illustration, we ran the algorithms under the unrealistic assumption that the ideal vocabulary size is already known; so that each model stopped the training routine once that size was reached. The numbers shown in Table \ref{tab:subword-stats} therefore report the performance of an ideal setting for the models, including information that would be unknown in a practical application. BPE and WordPiece reached that ideal vocabulary size only in 11 out of 100 cases (and even then did not provide an ideal morphological segmentation by any means). An accurate reduction of the vocabulary to its minimal representation was therefore rarely achieved.

The second problem results from the assumption that BPE and WordPiece lead to a monotonic increase of vocabulary size. This assumption does not hold true in the special case of numerals: Thanks to the high degree of compositionality, the smallest possible vocabulary size to construct the data is not necessarily the set of individual characters, but can be the set of employed morphemes instead. The Mandarin numerals for example only require 10 morphemes to construct numerals up to 40, while 19 distinct segments can be found in these forms. Depending on the complexity of a language's morphology and phonology, the monotonicity assumption might be violated, and the vocabulary size might \textit{decrease} for a while. However, this is not necessarily the case, as in more opaque languages like Hindi, the vocabulary size still increases monotonically.


\section{Discussion and Conclusion}\label{sec:5}

In this study, we have demonstrated an efficient, transparent, and robust workflow for the annotation and analysis of numeral systems. The workflow features a detailed annotation scheme for shared morphemes across word forms, accounts for potential allomorphy, and can be carried out in a computer-assisted manner, using a web-based annotation tool. As a result, we presented a small sample of annotated numeral systems from 25 typologically diverse languages from Eurasia and South America. 
We used this sample to evaluate how well unsupervised methods for automated morpheme segmentation work in extremely low-resource scenarios with an extraordinarily strong morphological signal. The results suggest that the major error source of these models is allomorphy. When this factor is accounted for, rather satisfactory morphological analyses can be inferred automatically. For future research on morpheme segmentation in low-resource scenarios, the handling of allomorphy will therefore be crucial.

Several statistical measures of numeral systems introduced here confirm intuitive correlations, such that smaller morpheme inventories necessarily entail a higher expressivity 
of the individual morphemes. It remains unclear, however, if a measure of morphological complexity can be inferred from our measures, since information-theoretic approaches that have been proposed to measure morphological complexity on corpus data do not convey any useful information about the morphological structure of numeral systems. 
 
We conclude that due to their high degree of compositionality, numerals serve as an ideal controlled sample for developing and testing the annotation and inference of morphological structures in multilingual wordlists. In the future, we hope to further expand our sample of numeral systems and test more methods for automated morpheme segmentation.

\section*{Supplementary Material}


All data and code underlying this study are made available within the Open Science Framework under \url{https://osf.io/786v9/?view_only=2f1f16e92abc46f7af5a8404448cb7b9}.

\section*{Limitations}

The annotation of word forms that etymologically share the same origin, but have diverged over a substantial amount of time, is not always clear and can be ambiguous. For example, consider Spanish \textit{once} (11): There is no transparent, synchronous pattern that would combine \textit{uno} (1) and \textit{diez} (10) to yield this form. However, we know that this was historically the case, as proven by Latin \textit{undecim}, which is a clear compound from \textit{un-} (1) and \textit{decem} (10). In Italian, this compounding strategy is still transparently visible (\textit{un-} + \textit{dieci} = \textit{ùndici}). Arguably, this lexical motivation is still transparent enough in Italian to annotate it as dimorphemic form, but not in Spanish (even though the etymology and the time depth is identical). A similar case can be observed for the Gaelic languages, where the suffix for deriving tens (Irish: \textit{déag}; Scottish: \textit{deug}) is clearly related to the word for ten (\textit{deich} in both languages), but the exact historical connection is unclear (\citealp[93-94]{matasovic2009etymological}; \citealp[130]{macbain1911etymological}).

Due to its relatively small size of 25 languages, the patterns observed in the data might not reflect universal patterns, especially considering the choice of languages. While we tried to include typologically diverse languages, we are aware that our sample is heavily biased towards Indo-European and Sino-Tibetan languages, and that the macroareas of North America, Africa, and Papunesia are not represented at all.

We furthermore observe a heavy bias towards decimal systems, and even those systems that are not primarily decimal contain some decimal structures. It is therefore impossible to systematically analyze different numeral bases beyond some impressionistic analyses. Finally, it remains an open question if (and how) the morphological complexity of a numeral system or a language in general can be measured.

\section*{Acknowledgments}

This project was supported by the ERC Consolidator Grant ProduSemy (AR, LC, AK, KB, DS, JML; Grant No. 101044282, see \url{https://doi.org/10.3030/101044282}), the ERC Synergy Grant QUANTA (CR; Grant No. 951388, see \url{https://doi.org/10.3030/951388}), and the Charles University (AS; project GA UK No. 101924, see \url{https://ufal.mff.cuni.cz/node/2690}). Views and opinions expressed are however those of the author(s) only and do not necessarily reflect those of the European Union or the European Research Council Executive Agency (nor any other funding agencies involved). Neither the European Union nor the granting authority can be held responsible for them.

\bibliography{custom}

\begin{thebibliography}{57}
\providecommand{\natexlab}[1]{#1}

\bibitem[{Batsuren et~al.(2024)Batsuren, Vylomova, Dankers, Delgerbaatar, Uzan, Pinter, and Bella}]{batsuren2024evaluating}
Khuyagbaatar Batsuren, Ekaterina Vylomova, Verna Dankers, Tsetsuukhei Delgerbaatar, Omri Uzan, Yuval Pinter, and G{\'a}bor Bella. 2024.
\newblock \href {https://doi.org/10.48550/arXiv.2404.13292} {Evaluating subword tokenization: Alien subword composition and oov generalization challenge}.
\newblock \emph{arXiv preprint arXiv:2404.13292}.

\bibitem[{Bender and Beller(2012)}]{Bender2012}
Andrea Bender and Sieghard Beller. 2012.
\newblock \href {https://doi.org/10.1016/j.cognition.2012.05.005} {Nature and culture of finger counting: Diversity and representational effects of an embodied cognitive tool}.
\newblock \emph{Cognition}, 124(2):156--182.

\bibitem[{Bentz et~al.(2017)Bentz, Alikaniotis, Cysouw, and Ferrer-i{-}Cancho}]{bentz2017entropy}
Christian Bentz, Dimitrios Alikaniotis, Michael Cysouw, and Ramon Ferrer-i{-}Cancho. 2017.
\newblock \href {https://doi.org/10.3390/e19060275} {The entropy of words -- learnability and expressivity across more than 1000 languages}.
\newblock \emph{Entropy}, 19(6).

\bibitem[{Brandt~Corstius(1968)}]{Corstius1968}
Hugo Brandt~Corstius, editor. 1968.
\newblock \emph{Grammars for number words}.
\newblock Reidel, Dordrecht.

\bibitem[{Brid et~al.(2022)Brid, Messineo, and List}]{Brid2022b}
Nicolás Brid, Cristina Messineo, and Johann-Mattis List. 2022.
\newblock \href {https://doi.org/10.12688/openreseurope.14922.2} {A comparative wordlist for the languages of the gran chaco, south america [version 2; peer review: 2 approved]}.
\newblock \emph{Open Research Europe}, 2(90):1--17.

\bibitem[{Chan(2024)}]{Chan2024}
Eugene Chan. 2024.
\newblock Numeral systems of the world’s languages.
\newblock \url{https://lingweb.eva.mpg.de/channumerals/}.
\newblock Version updated on February 18, 2024.

\bibitem[{{\c{C}}{\"o}ltekin and Rama(2023)}]{coltekin2022complexity}
{\c{C}}a{\u{g}}r{\i} {\c{C}}{\"o}ltekin and Taraka Rama. 2023.
\newblock \href {https://doi.org/doi:10.1515/lingvan-2021-0007} {What do complexity measures measure? {C}orrelating and validating corpus-based measures of morphological complexity}.
\newblock \emph{Linguistics Vanguard}, 9:27--43.

\bibitem[{Comrie(1997)}]{Comrie1997}
Bernard Comrie. 1997.
\newblock Some problems in the theory and typology of numeral systems.
\newblock In Bohumil Palek, editor, \emph{Proceedings of LP’96. Typology: Prototypes, item orderings, and universals. Proceedings of the conference held in Prague August 20–22, 1996}, pages 41--56. Charles University Press, Prague.

\bibitem[{Comrie(2005)}]{Comrie2005}
Bernard Comrie. 2005.
\newblock Endangered numeral systems.
\newblock In Jan Wohlgemut and Tyro Dirksmeyer, editors, \emph{Bedrohte Vielfalt Aspekte des Sprach(en)tods: Aspects of language death}, pages 203--230. Weissensee, Berlin.

\bibitem[{Comrie(2020)}]{Comrie2020}
Bernard Comrie. 2020.
\newblock \href {https://doi.org/10.22425/jul.2020.21.2.43} {Revisiting {Greenberg}’s ``{G}eneralizations about numeral systems'' (1978)}.
\newblock \emph{Journal of Universal Language}, 21(2):43–84.

\bibitem[{Creutz and Lagus(2002)}]{creutz2002unsupervised}
Mathias Creutz and Krista Lagus. 2002.
\newblock \href {https://doi.org/10.3115/1118647.1118650} {Unsupervised discovery of morphemes}.
\newblock In \emph{Proceedings of the {ACL}-02 Workshop on Morphological and Phonological Learning}, pages 21--30. Association for Computational Linguistics.

\bibitem[{Creutz and Lagus(2005)}]{creutz2005unsupervised}
Mathias Creutz and Krista Lagus. 2005.
\newblock \href {https://tuhat.helsinki.fi/ws/portalfiles/portal/62462066/Creutz05tr.pdf} {\emph{Unsupervised morpheme segmentation and morphology induction from text corpora using Morfessor 1.0}}.
\newblock Helsinki University of Technology Helsinki.

\bibitem[{Dixon(2012)}]{Dixon2012}
R.~M.~W. Dixon. 2012.
\newblock \emph{Basic linguistic theory, Vol. 3: Further grammatical topics}.
\newblock Oxford University Press, Oxford.

\bibitem[{Forkel(2024)}]{CLDFViz}
Robert Forkel. 2024.
\newblock \href {https://pypi.org/project/cldfviz} {\emph{{CLDFViz. A Python library providing tools to visualize data from CLDF datasets [Software Library, Version 1.3.0]}}}.
\newblock Zenodo, Geneva.

\bibitem[{Forkel and List(2020)}]{Forkel2020}
Robert Forkel and Johann-Mattis List. 2020.
\newblock \href {https://aclanthology.org/2020.lrec-1.864/} {{CLDFB}ench. {G}ive your cross-linguistic data a lift}.
\newblock In \emph{{P}roceedings of the {T}welfth {I}nternational {C}onference on {L}anguage {R}esources and {E}valuation}, page 6997‑7004, Luxembourg. European Language Resources Association (ELRA).

\bibitem[{Forkel and List(2024)}]{forkel2024linse}
Robert Forkel and Johann-Mattis List. 2024.
\newblock \href {https://doi.org/10.15475/calcip.2024.1.3} {A new {P}ython library for the manipulation and annotation of linguistic sequences}.
\newblock \emph{Computer-Assisted Language Comparison in Practice}, 7(1):17–23.

\bibitem[{Forkel et~al.(2018)Forkel, List, Greenhill, Rzymski, Bank, Cysouw, Hammarstr{\"o}m, Haspelmath, Kaiping, and Gray}]{forkel2018cross}
Robert Forkel, Johann-Mattis List, Simon~J. Greenhill, Christoph Rzymski, Sebastian Bank, Michael Cysouw, Harald Hammarstr{\"o}m, Martin Haspelmath, Gereon~A. Kaiping, and Russell~D. Gray. 2018.
\newblock \href {https://doi.org/10.1038/sdata.2018.205} {Cross-linguistic data formats, advancing data sharing and re-use in comparative linguistics}.
\newblock \emph{Scientific data}, 5(1):1--10.

\bibitem[{Gage(1994)}]{gage1994new}
Philip Gage. 1994.
\newblock A new algorithm for data compression.
\newblock \emph{C Users Journal}, 12(2):23--38.

\bibitem[{Goldsmith(2001)}]{goldsmith2001unsupervised}
John~A. Goldsmith. 2001.
\newblock \href {https://doi.org/10.1162/089120101750300490} {Unsupervised learning of the morphology of a natural language}.
\newblock \emph{Computational Linguistics}, 27(2):153--198.

\bibitem[{Goldsmith et~al.(2017)Goldsmith, Lee, and Xanthos}]{goldsmith2017computational}
John~A. Goldsmith, Jackson~L. Lee, and Aris Xanthos. 2017.
\newblock \href {https://doi.org/10.1146/annurev-linguistics-011516-034017} {Computational learning of morphology}.
\newblock \emph{Annual Review of Linguistics}, 3:85--106.

\bibitem[{Gower(2021)}]{CSVW}
Robin Gower. 2021.
\newblock \href {https://csvw.org} {\emph{CSV on the Web}}.
\newblock Swirrl, Stirling.

\bibitem[{Greenberg(1978)}]{Greenberg1978}
Joseph~H. Greenberg. 1978.
\newblock Generalizations about numeral systems.
\newblock In Joseph~H. Greenberg, editor, \emph{Universals of Human Language, Vol. 3: Word structure}, pages 249--295. Stanford University Press, Stanford.

\bibitem[{Hafer and Weiss(1974)}]{hafer1974word}
Margaret~A. Hafer and Stephen~F. Weiss. 1974.
\newblock \href {https://doi.org/10.1016/0020-0271(74)90044-8} {Word segmentation by letter successor varieties}.
\newblock \emph{Information storage and retrieval}, 10(11-12):371--385.

\bibitem[{Hammarstr{\"o}m(2008)}]{hammarstrom2008complexity}
Harald Hammarstr{\"o}m. 2008.
\newblock \href {https://doi.org/10.1075/slcs.94} {Complexity in numeral systems with an investigation into pidgins and creoles}.
\newblock In Matti Miestamo, Kaius Sinnemäki, and Fred Karlsson, editors, \emph{Language Complexity. Typology, contact, change}, volume~94 of \emph{Studies in Language Companion Series}. John Benjamins Publishing Company.

\bibitem[{Hammarstr{\"o}m(2009)}]{hammarstrom2009unsupervised}
Harald Hammarstr{\"o}m. 2009.
\newblock \emph{Unsupervised Learning of Morphology and the Languages of the World}.
\newblock Ph.D. thesis, Chalmers University of Technology and University of Gothenburg.

\bibitem[{Hammarstr{\"o}m(2010)}]{Hammarstrom2010}
Harald Hammarstr{\"o}m. 2010.
\newblock \href {https://doi.org/10.1515/9783110220933} {Rarities in numeral systems}.
\newblock In Jan Wohlgemuth and Michael Cysouw, editors, \emph{Rethinking universals: How rarities affect linguistic theory}, volume~45 of \emph{Empirical Approaches to Language Typology}, pages 11--60. Mouton de Gruyter, Berlin.

\bibitem[{Hammarstr{\"o}m and Borin(2011)}]{hammarstrom2011ulm}
Harald Hammarstr{\"o}m and Lars Borin. 2011.
\newblock \href {https://doi.org/10.1162/coli_a_00050} {Unsupervised learning of morphology}.
\newblock \emph{Computational Linguistics}, 37(2):309--350.

\bibitem[{Hammarström et~al.(2024)Hammarström, Haspelmath, Forkel, and Bank}]{Glottolog}
Harald Hammarström, Martin Haspelmath, Robert Forkel, and Sebastian Bank. 2024.
\newblock \href {https://arxiv.org/abs/https://glottolog.org} {\emph{{G}lottolog [{Dataset, Version 5.1}]}}.
\newblock Max Planck Institute for Evolutionary Anthropology, Leipzig.

\bibitem[{Harris(1955)}]{harris1955from}
Zellig~S Harris. 1955.
\newblock \href {https://doi.org/10.2307/411036} {From phoneme to morpheme}.
\newblock \emph{Language}, 31(2):190.

\bibitem[{Heine(1997)}]{Heine1997}
Bernd Heine. 1997.
\newblock \href {https://doi.org/10.1093/oso/9780195102512.001.0001} {\emph{Cognitive foundations of grammar}}.
\newblock Oxford University Press, New York and Oxford.

\bibitem[{Herv{\'a}s~y Panduro(1786)}]{HervasyPanduro1786}
Lorenzo Herv{\'a}s~y Panduro. 1786.
\newblock \emph{Aritmetica delle nazioni e divisione del tempo fra l’Orientali}.
\newblock Gregorio Biasini, Cesena.

\bibitem[{Hill and List(2017)}]{Hill2017a}
Nathan~W. Hill and Johann-Mattis List. 2017.
\newblock \href {https://doi.org/10.1515/yplm-2017-0003} {{C}hallenges of annotation and analysis in computer-assisted language comparison: {A} case study on {B}urmish languages}.
\newblock \emph{Yearbook of the Poznań Linguistic Meeting}, 3(1):47–76.

\bibitem[{Hurford(1975)}]{Hurford1975}
James~R. Hurford. 1975.
\newblock \emph{The linguistic theory of numerals}.
\newblock Cambridge University Press, Cambridge.

\bibitem[{Kudo(2018)}]{kudo2018subword}
Taku Kudo. 2018.
\newblock \href {https://doi.org/10.18653/v1/P18-1007} {Subword regularization: Improving neural network translation models with multiple subword candidates}.
\newblock In \emph{Proceedings of the 56th Annual Meeting of the Association for Computational Linguistics (Volume 1: Long Papers)}, pages 66--75, Melbourne, Australia. Association for Computational Linguistics.

\bibitem[{List(2014)}]{List2014d}
Johann-Mattis List. 2014.
\newblock \href {https://doi.org/10.1515/9783110720082} {\emph{{S}equence comparison in historical linguistics}}.
\newblock Düsseldorf University Press, Düsseldorf.

\bibitem[{List(2017)}]{List2017d}
Johann-Mattis List. 2017.
\newblock \href {https://aclanthology.org/E17-3003} {{A} web-based interactive tool for creating, inspecting, editing, and publishing etymological datasets}.
\newblock In \emph{{P}roceedings of the 15th {C}onference of the {E}uropean {C}hapter of the {A}ssociation for {C}omputational {L}inguistics. {S}ystem {D}emonstrations}, pages 9--12, Valencia. Association for Computational Linguistics.

\bibitem[{List(2023)}]{List2023a}
Johann-Mattis List. 2023.
\newblock \href {https://doi.org/10.3389/fpsyg.2023.1156540} {Inference of partial colexifications from multilingual wordlists}.
\newblock \emph{Frontiers in Psychology}, 14(1156540):1--10.

\bibitem[{List(2024)}]{List2024fc}
Johann-Mattis List. 2024.
\newblock \href {https://doi.org/10.17613/zfwr-sn25} {Productive signs: A computer-assisted analysis of evolutionary, typological, and cognitive dimensions of word families}.
\newblock In \emph{International Conference of Linguists}, pages 1--12. Brill, Pozna\'n.

\bibitem[{List et~al.(2021)List, Anderson, Tresoldi, and Forkel}]{CLTS}
Johann-Mattis List, Cormac Anderson, Tiago Tresoldi, and Robert Forkel. 2021.
\newblock \href {https://arxiv.org/abs/https://clts.clld.org} {\emph{{C}ross-{L}inguistic {T}ranscription {S}ystems. [Dataset, {V}ersion 2.1.0]}}.
\newblock Max Planck Institute for the Science of Human History, Jena.

\bibitem[{List et~al.(2022)List, Forkel, Greenhill, Rzymski, Englisch, and Gray}]{List2022e}
Johann-Mattis List, Robert Forkel, Simon~J. Greenhill, Christoph Rzymski, Johannes Englisch, and Russell~D. Gray. 2022.
\newblock \href {https://doi.org/10.1038/s41597-022-01432-0} {Lexibank, a public repository of standardized wordlists with computed phonological and lexical features}.
\newblock \emph{Scientific Data}, 9(316):1--31.

\bibitem[{List et~al.(2016)List, Lopez, and Bapteste}]{List2016g}
Johann-Mattis List, Philippe Lopez, and Eric Bapteste. 2016.
\newblock \href {http://anthology.aclweb.org/P16-2097} {{U}sing sequence similarity networks to identify partial cognates in multilingual wordlists}.
\newblock In \emph{{P}roceedings of the {A}ssociation of {C}omputational {L}inguistics 2016 ({V}olume 2: {S}hort {P}apers)}, pages 599--605, Berlin. Association of Computational Linguistics.

\bibitem[{List et~al.(2025{\natexlab{a}})List, Tjuka, Blum, Kučerová, Barrientos~Ugarte, Rzymski, Greenhill, and Forkel}]{Concepticon}
Johann-Mattis List, Annika Tjuka, Frederic Blum, Alžběta Kučerová, Carlos Barrientos~Ugarte, Christoph Rzymski, Simon~J. Greenhill, and Robert Forkel. 2025{\natexlab{a}}.
\newblock \href {https://arxiv.org/abs/https://concepticon.clld.org/} {\emph{{CLLD Concepticon} [{Dataset, Version 3.3.0}]}}.
\newblock Max Planck Institute for Evolutionary Anthropology, Leipzig.

\bibitem[{List and van Dam(2024)}]{List2024b}
Johann-Mattis List and Kellen~Parker van Dam. 2024.
\newblock \href {https://aclanthology.org/2024.lchange-1.1} {Computer-assisted language comparison with {EDICTOR} 3 [invited paper]}.
\newblock In \emph{Proceedings of the 5th Workshop on Computational Approaches to Historical Language Change}, pages 1--11, Bangkok, Thailand. Association for Computational Linguistics.

\bibitem[{List et~al.(2025{\natexlab{b}})List, {van Dam}, and Blum}]{EDICTOR}
Johann-Mattis List, Kellen~Parker {van Dam}, and Frederic Blum. 2025{\natexlab{b}}.
\newblock \href {https://edictor.org} {\emph{EDICTOR 3. An Interactive Tool for Computer-Assisted Language Comparison [Software Tool, Version 3.1]}}.
\newblock MCL Chair at the University of Passau, Passau.

\bibitem[{MacBain(1911)}]{macbain1911etymological}
Alexander MacBain. 1911.
\newblock \emph{An etymological dictionary of the Gaelic language}.
\newblock Eneas Mackay, Stirling.

\bibitem[{Matasovi\'c(2009)}]{matasovic2009etymological}
Ranko Matasovi\'c. 2009.
\newblock \emph{Etymological dictionary of Proto-Celtic}.
\newblock Brill, Leiden, Boston.

\bibitem[{Moravcsik(2017)}]{Moravcsik2017}
Edith~A. Moravcsik. 2017.
\newblock \href {https://doi.org/10.1017/9781316135716} {Number}.
\newblock In Alexandra~Y. Aikhenvald and R.~M.~W. Dixon, editors, \emph{The Cambridge handbook of linguistic typology}, pages 440--476. Cambridge University Press, Cambridge.

\bibitem[{Pulini and List(2024)}]{Pulini2024}
Michele Pulini and Johann-Mattis List. 2024.
\newblock \href {https://doi.org/10.30184/BCL.202410_17.0004} {Finding language-internal cognates in {O}ld {C}hinese}.
\newblock \emph{Bulletin of Chinese Linguistics}, 17(1):53--72.

\bibitem[{Rubehn and List(2025)}]{Rubehn2025PREPRINT}
Arne Rubehn and Johann-Mattis List. 2025.
\newblock \href {https://doi.org/10.48550/arXiv.2502.09743} {Partial colexifications improve concept embeddings}.
\newblock \emph{arXiv}, 2502(09743):1--15.

\bibitem[{Rzymski et~al.(2020)Rzymski, Tresoldi, Greenhill, Wu, Schweikhard, Koptjevskaja-Tamm, Gast, Bodt, Hantgan, Kaiping, Chang, Lai, Morozova, Arjava, Hübler, Koile, Pepper, Proos, Epps, Blanco, Hundt, Monakhov, Pianykh, Ramesh, Gray, Forkel, and List}]{Rzymski2020}
Christoph Rzymski, Tiago Tresoldi, Simon Greenhill, Mei-Shin Wu, Nathanael~E. Schweikhard, Maria Koptjevskaja-Tamm, Volker Gast, Timotheus~A. Bodt, Abbie Hantgan, Gereon~A. Kaiping, Sophie Chang, Yunfan Lai, Natalia Morozova, Heini Arjava, Nataliia Hübler, Ezequiel Koile, Steve Pepper, Mariann Proos, Briana~Van Epps, Ingrid Blanco, Carolin Hundt, Sergei Monakhov, Kristina Pianykh, Sallona Ramesh, Russell~D. Gray, Robert Forkel, and Johann-Mattis List. 2020.
\newblock \href {https://doi.org/10.1038/s41597-019-0341-x} {{T}he {D}atabase of {C}ross-{L}inguistic {C}olexifications, reproducible analysis of cross-linguistic polysemies}.
\newblock \emph{Scientific Data}, 7(13):1--12.

\bibitem[{Schuster and Nakajima(2012)}]{schuster2012japanese}
Mike Schuster and Kaisuke Nakajima. 2012.
\newblock \href {https://doi.org/10.1109/ICASSP.2012.6289079} {Japanese and {K}orean voice search}.
\newblock In \emph{2012 IEEE International Conference on Acoustics, Speech and Signal Processing (ICASSP)}, pages 5149--5152, Kyoto, Japan. IEEE.

\bibitem[{Schweikhard and List(2020)}]{Schweikhard2020}
Nathanael~E. Schweikhard and Johann-Mattis List. 2020.
\newblock \href {http://www.skase.sk/Volumes/JTL43/index.html} {Developing an annotation framework for word formation processes in comparative linguistics}.
\newblock \emph{SKASE Journal of Theoretical Linguistics}, 17(1):2--26.

\bibitem[{Sennrich et~al.(2016)Sennrich, Haddow, and Birch}]{sennrich2016neural}
Rico Sennrich, Barry Haddow, and Alexandra Birch. 2016.
\newblock \href {https://doi.org/10.18653/v1/P16-1162} {Neural machine translation of rare words with subword units}.
\newblock In \emph{Proceedings of the 54th Annual Meeting of the Association for Computational Linguistics (Volume 1: Long Papers)}, pages 1715--1725, Berlin, Germany. Association for Computational Linguistics.

\bibitem[{Skirg{\aa}rd et~al.(2023)Skirg{\aa}rd, Haynie, Blasi, Hammarstr\"{o}m, Collins, Latarche, Lesage, Weber, Witzlack-Makarevich, Passmore, Chira, Maurits, Dinnage, Dunn, Reesink, Singer, Bowern, Epps, Hill, Vesakoski, Robbeets, Abbas, Auer, Bakker, Barbos, Borges, Danielsen, Dorenbusch, Dorn, Elliott, Falcone, Fischer, Ate, Gibson, G\"{o}bel, Goodall, Gruner, Harvey, Hayes, Heer, Miranda, H\"{u}bler, Huntington-Rainey, Ivani, Johns, Just, Kashima, Kipf, Klingenberg, K\"{o}nig, Koti, Kowalik, Krasnoukhova, Lindvall, Lorenzen, Lutzenberger, Martins, German, van~der Meer, Samam{\'{e}}, M\"{u}ller, Muradoglu, Neely, Nickel, Norvik, Oluoch, Peacock, Pearey, Peck, Petit, Pieper, Poblete, Prestipino, Raabe, Raja, Reimringer, Rey, Rizaew, Ruppert, Salmon, Sammet, Schembri, Schlabbach, Schmidt, Skilton, Smith, de~Sousa, Sverredal, Valle, Vera, Vo{\ss}, Witte, Wu, Yam, Ye, Yong, Yuditha, Zariquiey, Forkel, Evans, Levinson, Haspelmath, Greenhill, Atkinson, and Gray}]{Skirgard2023}
Hedvig Skirg{\aa}rd, Hannah~J. Haynie, Dami{\'{a}}n~E. Blasi, Harald Hammarstr\"{o}m, Jeremy Collins, Jay~J. Latarche, Jakob Lesage, Tobias Weber, Alena Witzlack-Makarevich, Sam Passmore, Angela Chira, Luke Maurits, Russell Dinnage, Michael Dunn, Ger Reesink, Ruth Singer, Claire Bowern, Patience Epps, Jane Hill, Outi Vesakoski, Martine Robbeets, Noor~Karolin Abbas, Daniel Auer, Nancy~A. Bakker, Giulia Barbos, Robert~D. Borges, Swintha Danielsen, Luise Dorenbusch, Ella Dorn, John Elliott, Giada Falcone, Jana Fischer, Yustinus~Ghanggo Ate, Hannah Gibson, Hans-Philipp G\"{o}bel, Jemima~A. Goodall, Victoria Gruner, Andrew Harvey, Rebekah Hayes, Leonard Heer, Roberto E.~Herrera Miranda, Nataliia H\"{u}bler, Biu Huntington-Rainey, Jessica~K. Ivani, Marilen Johns, Erika Just, Eri Kashima, Carolina Kipf, Janina~V. Klingenberg, Nikita K\"{o}nig, Aikaterina Koti, Richard G.~A. Kowalik, Olga Krasnoukhova, Nora L.~M. Lindvall, Mandy Lorenzen, Hannah Lutzenberger, T{\^{a}}nia R.~A. Martins, Celia~Mata German, Suzanne
  van~der Meer, Jaime~Montoya Samam{\'{e}}, Michael M\"{u}ller, Saliha Muradoglu, Kelsey Neely, Johanna Nickel, Miina Norvik, Cheryl~Akinyi Oluoch, Jesse Peacock, India O.~C. Pearey, Naomi Peck, Stephanie Petit, S\"{o}ren Pieper, Mariana Poblete, Daniel Prestipino, Linda Raabe, Amna Raja, Janis Reimringer, Sydney~C. Rey, Julia Rizaew, Eloisa Ruppert, Kim~K. Salmon, Jill Sammet, Rhiannon Schembri, Lars Schlabbach, Frederick W.~P. Schmidt, Amalia Skilton, Wikaliler~Daniel Smith, Hil{\'{a}}rio de~Sousa, Kristin Sverredal, Daniel Valle, Javier Vera, Judith Vo{\ss}, Tim Witte, Henry Wu, Stephanie Yam, Jingting Ye, Maisie Yong, Tessa Yuditha, Roberto Zariquiey, Robert Forkel, Nicholas Evans, Stephen~C. Levinson, Martin Haspelmath, Simon~J. Greenhill, Quentin~D. Atkinson, and Russell~D. Gray. 2023.
\newblock \href {https://doi.org/10.1126/sciadv.adg6175} {Grambank reveals the importance of genealogical constraints on linguistic diversity and highlights the impact of language loss}.
\newblock \emph{Science Advances}, 9(16).

\bibitem[{Spiegler and Monson(2010)}]{spiegler2010emma}
Sebastian Spiegler and Christian Monson. 2010.
\newblock \href {https://aclanthology.org/C10-1116/} {{EMMA}: A novel evaluation metric for morphological analysis}.
\newblock In \emph{Proceedings of the 23rd International Conference on Computational Linguistics (Coling 2010)}, pages 1029--1037, Beijing, China. Coling 2010 Organizing Committee.

\bibitem[{Virpioja et~al.(2013)Virpioja, Smit, Gr{\"o}nroos, and Kurimo}]{virpioja2013morfessor}
Sami Virpioja, Peter Smit, Stig-Arne Gr{\"o}nroos, and Mikko Kurimo. 2013.
\newblock \href {https://urn.fi/URN:ISBN:978-952-60-5501-5} {Morfessor 2.0: {P}ython implementation and extensions for {M}orfessor {B}aseline}.
\newblock In \emph{Aalto University publication series SCIENCE + TECHNOLOGY}, 25. Aalto University, Helsinki, Finland.

\bibitem[{Virpioja et~al.(2011)Virpioja, Turunen, Spiegler, Kohonen, and Kurimo}]{virpioja2011empirical}
Sami Virpioja, Ville~T Turunen, Sebastian Spiegler, Oskar Kohonen, and Mikko Kurimo. 2011.
\newblock \href {https://aclanthology.org/2011.tal-2.3/} {Empirical comparison of evaluation methods for unsupervised learning of morphology}.
\newblock \emph{Traitement Automatique des Langues}, 52(2):45--90.

\end{thebibliography}

\appendix
\onecolumn

\newpage
\section{Statistics for Individual Languages}
\label{sec:appendix}

\begin{table*}[!ht]
\centering
\begin{tabular}{@{}llllllll@{}}
\toprule
\textbf{Language}           & \textbf{Morph.} & \textbf{Expressivity} & \textbf{Opacity} & \textbf{Length} & \textbf{Morfessor} & \textbf{LSPE} & \textbf{Affix} \\ \midrule
Maltese            &   25 / 12 & 4.24 / 8.83 & 2.08 & 2.65 & 0.71 / 0.62 & 0.64 / 0.84 & 0.71 / 0.84     \\
Mapudungun         &   10 / 10 & 8.80 / 8.80 & 1.00 & 2.20 & 1.00 / 1.00 & 1.00 / 1.00 & 1.00 / 1.00   \\
Wayuu              &  18 / 15 & 7.26 / 8.71 & 1.20 & 3.19 & 0.90 / 0.90 & 0.89 / 0.95 & 0.70 / 0.70   \\
Aymara             &  12 / 10 & 10.58 / 12.70 & 1.20 & 3.17 & 0.94 / 1.00 & 0.67 / 0.65 & 0.73 / 0.74   \\
Telugu             &   27 / 12 & 3.26 / 7.33 & 2.25 & 2.20 & 0.53 / 0.88 & 0.35 / 0.72 & 0.56 / 0.99   \\
Czech              &   17 / 11 & 5.18 / 8.00 & 1.55 & 2.20 & 0.73 / 1.00 & 0.86 / 0.82 & 0.95 / 1.00   \\
Russian            &   17 / 12 & 5.65 / 8.00 & 1.42 & 2.40 & 0.62 / 0.91 & 0.69 / 0.81 & 0.83 / 0.97   \\
Irish              &   22 / 14 & 4.39 / 6.89 & 1.57 & 2.41 & 0.61 / 0.89 & 0.58 / 0.63 & 0.75 / 0.99  \\
Scottish G.    &    17 / 13 & 6.94 / 9.08 & 1.31 & 2.95 & 0.61 / 0.66 & 0.76 / 0.90 & 0.61 / 0.81  \\
German             &    20 / 15 & 5.20 / 6.93 & 1.33 & 2.60 & 0.81 / 0.81 & 0.65 / 0.95 & 0.80 / 0.83   \\
Assamese           &   41 / 20 & 1.66 / 3.40 & 2.05 & 1.70 & 0.64 / 1.00 & 0.55 / 0.79 & 0.60 / 0.88  \\
Hindi              &   48 / 17 & 1.40 / 3.94 & 2.82 & 1.68 & 0.42 / 1.00 & 0.46 / 0.95 & 0.43 / 1.00   \\
Sanskrit           &   29 / 13 & 3.14 / 7.00 & 2.23 & 2.53 & 0.66 / 0.70 & 0.55 / 0.53 & 0.45 / 0.75   \\
French             &    24 / 19 & 3.08 / 3.89 & 1.26 & 1.85 & 0.75 / 0.79 & 0.73 / 0.80 & 0.67 / 1.00     \\
Italian            &   27 / 14 & 3.07 / 5.93 & 1.93 & 2.08 & 0.75 / 0.82 & 0.67 / 0.77 & 0.79 / 0.96   \\
Latin              &   23 / 15 & 4.00 / 6.13 & 1.53 & 2.30 & 0.73 / 0.82 & 0.71 / 0.79 & 0.65 / 0.86    \\
Spanish            &   25 / 19 & 3.92 / 5.16 & 1.32 & 2.45 & 0.55 / 0.87 & 0.82 / 0.89 & 0.73 / 0.97   \\
Cavineña           &   13 / 10 & 10.46 / 13.60 & 1.30 & 3.40 & 0.93 / 1.00 & 0.46 / 0.57 & 0.67 / 0.68   \\
H. Quechua   &       10 / 10 & 8.80 / 8.80 & 1.00 & 2.20 & 1.00 / 1.00 & 0.89 / 0.89 & 1.00 / 1.00   \\
Lamjung Y.      &     31 / 11 & 3.16 / 8.91 & 2.82 & 2.45 & 0.65 / 0.89 & 0.80 / 0.88 & 0.61 / 0.89     \\
Uipo (M.)    &      18 / 18 & 8.50 / 8.50 & 1.00 & 3.83 & 0.40 / 0.40 & 0.93 / 0.93 & 0.40 / 0.40   \\
Makyam     &    27 / 18 & 4.81 / 7.22 & 1.50 & 3.25 & 0.70 / 0.85 & 0.70 / 0.71 & 0.45 / 0.77    \\
Mandarin           &   10 / 10 & 8.80 / 8.80 & 1.00 & 2.20 & 1.00 / 1.00 & 1.00 / 1.00 & 0.96 / 0.96  \\
Shanghainese       &   12 / 11 & 6.50 / 7.09 & 1.09 & 1.95 & 1.00 / 1.00 & 0.87 / 0.87 & 1.00 / 1.00   \\
Par. Guaraní &      11 / 7 & 9.55 / 15.00 & 1.57 & 2.62 & 0.91 / 1.00 & 0.89 / 1.00 & 0.99 / 1.00   \\ \bottomrule
\end{tabular}
\caption{Overview of statistics about the different numeral systems for each individual language. Whenever two values are given, the left refers to the surface forms, and the right to the underlying form. Morph. indicates the number of distinct morph(eme)s in the given language. The three rightmost columns indicate the performance of automated morpheme segmentation models in terms of F$_1$.}
\label{tab:stats-all-langs}
\end{table*}

\end{document}